\pdfoutput=1
\documentclass[11pt]{article}
\usepackage{emnlp2021}
\usepackage{times}
\usepackage{latexsym}
\usepackage[T1]{fontenc}
\usepackage[utf8]{inputenc}
\usepackage{microtype}
\usepackage{graphicx}
\usepackage{amsmath,amssymb}
\DeclareMathOperator*{\E}{\mathbb{E}}
\DeclareMathOperator*{\argmin}{arg\,min}

\title{Open-domain clarification question generation without question examples}

\author{Julia White \\ Electrical Engineering \\  Stanford University \And  
Gabriel Poesia \\ Computer Science \\ Stanford University \And
Robert Hawkins \\ Psychology \\ Princeton University \AND
Dorsa Sadigh \\ Electrical Engineering, Computer Science \\ Stanford University\And
Noah Goodman \\ Computer Science, Psychology \\ Stanford University} 
  
\begin{document}
\maketitle
\begin{abstract}
An overarching goal of natural language processing is to enable machines to communicate seamlessly with humans. However, natural language can be ambiguous or unclear. In cases of uncertainty, humans engage in an interactive process known as repair: asking questions and seeking clarification until their uncertainty is resolved. We propose a framework for building a visually grounded question-asking model capable of producing polar (yes-no) clarification questions to resolve misunderstandings in dialogue. Our model uses an expected information gain objective to derive informative questions from an off-the-shelf image captioner without requiring any supervised question-answer data. We demonstrate our model's ability to pose questions that improve communicative success in a goal-oriented 20 questions game with synthetic and human answerers.
\end{abstract}

\section{Introduction}
Human-machine interaction relies on accurate transfer of knowledge from users. However, natural language input can be ambiguous or unclear, giving rise to uncertainty.
A fundamental aspect of human communication is collaborative grounding, or seeking and providing incremental evidence of mutual understanding through dialog. Specifically, humans can correct for uncertainty through cooperative repair \cite{clark1996,purver2001,arkel2020} which involves interactively asking questions and seeking clarification. Making and recovering from mistakes collaboratively through question-asking is a key ingredient in grounding meaning and therefore an important feature in dialog systems \cite{benotti2021}. 
In this work, we focus on the computational challenge of generating clarification questions in visually grounded human-machine interactions.

\begin{figure}[tbh]
\begin{center}
\includegraphics[width=7.25cm]{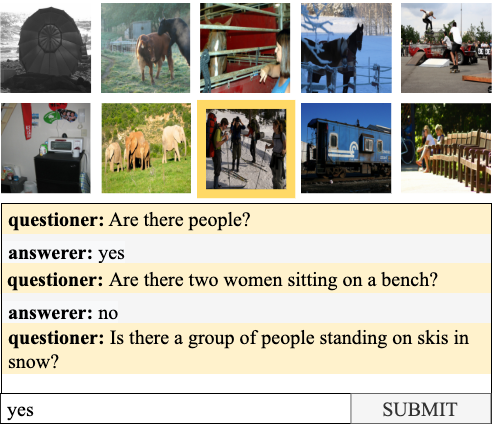}
\end{center}
\caption{Our model takes the role of questioner in a question-driven communication game where it must guess which image is being described by the answerer. The interaction ends with the model returning a guess for which image the answerer is referring to.}
\label{game}
\end{figure}

One popular approach is to train an end-to-end model to map visual and linguistic inputs directly to questions \cite{yao2018, das2017}.
This approach is heavily data-driven, requiring large annotated training sets of questions under different goals and contexts. 
Another approach has drawn from work on active learning and Optimal Experiment Design (OED) in cognitive science to \emph{search} for questions that are likely to maximize expected information gain from an imagined answerer \cite{wang2020,lee2018,misra2018,rao2018,rothe2017,kovashka2013}.
Much of this work has relied on large-scale question-answer datasets \cite{kumar2020,vries2017} for training or retrieval to propose candidate questions or evaluate their expected utility. 
Others, like \cite{yu2020}, derive questions from attribute annotations for domain-specific systems.

\begin{figure*} 
\begin{center}
\includegraphics[width=15cm]{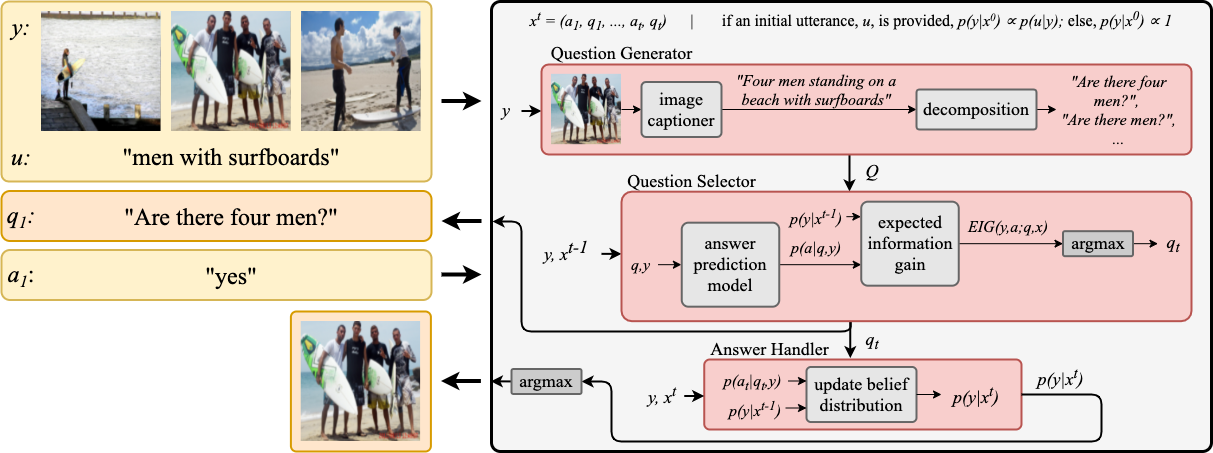}
\end{center}
\caption{A set of candidate questions are produced by our question generator, and then ranked according to their expected utility in the question selector module. After posing the highest-ranked question and receiving an answer, the belief distribution over images is updated in the answer handler module and these updated beliefs are then either used to guess the target image or are fed back to the question selection module for the process to be repeated.
}
\label{model}
\end{figure*}

In this paper, we address an open-domain setting where one cannot rely on an immediate grounding of the meaning of questions in the target domain (in contrast to end-to-end approaches, which assume examples of questions to train on, or semantic parsing approaches, which assume a logical form for questions). Our key contribution is a lightweight method to ground question semantics in the open image domain without observing question examples. Instead, our framework builds a visually grounded question-asking model from image captioning data, deriving question selection and belief updating without existing semantics. 
Our model generates candidate polar questions, arguably the most common form of clarification in dialogue \cite{stivers2010}, by applying rule-based linguistic transformations to the outputs of a pretrained image captioner. 
We then use self-supervision to train a response model that predicts the likelihood of different answers. 
Given these predictions, we estimate the expected information gain of each question and select the question with the highest utility. 
We demonstrate our method's ability to pose questions that improve communicative success in a question-driven communication game with synthetic and human answerers.

\section{20 Questions Task}
We study interactions between questioners and answerers in a visually grounded 20-questions paradigm (see Figure~\ref{game}).
Both agents are shown a set of $k$ images as a context ($k = 10$ in Figure~\ref{game}).
One of these images is privately indicated to the answerer as the \emph{target} (e.g., ~bottom row, center), but remains unknown to the questioner. 
The questioner's goal is therefore to select questions that allow them to identify this target based on responses from the answerer. 
After a maximum of 20 questions, the questioner must make a guess (i.e., a~$k$-way classification). 
This task can be viewed as the most straightforward extension of a signaling game \cite{lewis1969} to allow for interactive clarification and repair. 
To approximate the setting of natural ``clarification questions'' we also consider games that begin with a description of the target.
Critically, the appropriate question changes depending on the context of objects and previous information provided by the answerer. 

\section{Model} 
Our model (Figure~\ref{model}) maintains a belief distribution, $p(y|x^t)$, about which image $y$ in the set of images $Y$ is the target.
This distribution is conditioned on the history of the interaction, $x^t=(a_1,q_1,...,a_t,q_t)$, which includes all questions, $q$, and answers, $a$, exchanged up to the current step, $t$. 
Our model is defined in terms of three basic components.
At each interaction step, it must \emph{generate} a set of candidate questions, \emph{select} one of these candidates based on expected information gain and finally \emph{update} its beliefs based on the answer.

\textbf{Question Generator.}
To generate questions without question examples, we must derive suitable candidates using an alternative method.
Specifically, we suggest using a pre-trained image captioner to produce a list of candidate \emph{captions}, which can then be programmatically transformed into question form. 
We begin by producing a list of captions for each image $y\in Y$ and decomposing each of these captions into multiple polar questions according to a constituency parse, obtained using the Berkeley Neural Parser \cite{kitaev2019}. 
We then transform each noun phrase (NP) subtree in each caption's constituency tree into a polar question (`Are there <NP>?' with indefinite articles and plurality chosen for agreement).
Using this procedure, we generate an average of 10 candidate questions from each caption (see Appendix A for examples).

\textbf{Question Selector.}
To determine the most informative question at turn $t+1$, we estimate expected information gain, $EIG(y,a;q,x^t)$, for every question in the candidate set $Q$ (after a question is asked it is removed from the set).
EIG is defined as the change in entropy of the distribution over images after observing an answer $a \in A(q)$ to question $q$.
Because the initial entropy is the same for every question, maximizing the EIG is equivalent to minimizing the expected conditional entropy of the belief distribution under possible answers.
Because different answers are expected given different targets $y$, we marginalize over $a$ inside the entropy:
\vspace{-1em}
\begin{equation}
\argmin_{q\in Q} \E_{p(y | x^{t})}\E_{p(a|q,y)}\left[-\ln P(y | x^{t}, q, a)\right]
\end{equation}

The distribution
$p(a|q,y)$ represents predictions about how the answerer will respond to a question when $y$ is the target.
We do not have access to a ground-truth answerer model, so we amortize these predictions by training an answer classifier.
We introduced a self-supervision objective by either pairing target images with questions derived from them (`yes' answers) or with questions derived from other images (`no' answers).
It should be noted that this data-generation method may occasionally yield a false negative when, for a `no'-labelled question-image pair, a question is sampled that does coincidentally apply to the image; however, these samples represent a minority of the training data. 
We then trained a logistic classifier using cross-entropy loss on concatenated image and caption embeddings obtained from a CNN and RNN encoder, respectively.
This classifier yields a prediction of yes vs. no answers for any unseen pair $(y, q)$ with 94\% accuracy on held-out, manually-labelled datapoints. 

\textbf{Answer Handler.}
Finally, after obtaining an answer $a$, our model must update its beliefs for the subsequent time step (and anticipate this update for Eq.~1). 
The belief update is given by Bayes rule:
\begin{equation*}
\begin{split}
p(y|x^{t},q,a) \propto &\,\,
p(a|x^{t},q,y) p(q|x^{t},y)  p(y|x^{t})
\end{split}
\end{equation*}
The first term $p(a|x^{t},q,y)$ can be simplified to our amortized answer prediction model described above by assuming that the answer is independent of past interactions.
The second term is given by the deterministic question selector model described above.
The third term is given by the belief distribution on the previous time step. 
The initial belief distribution is either uniform, $p(y|x^0)\propto 1$, or, when an initial description $u$ is provided, it is proportional to the utterance likelihood under the captioning model, $p(y|x^0)\propto p(u|y)$.

\section{Experiments}

We evaluated our question-asking framework in grounded interactions with both synthetic and human answerers.

\begin{figure*}
\begin{center}
\includegraphics[width=15cm]{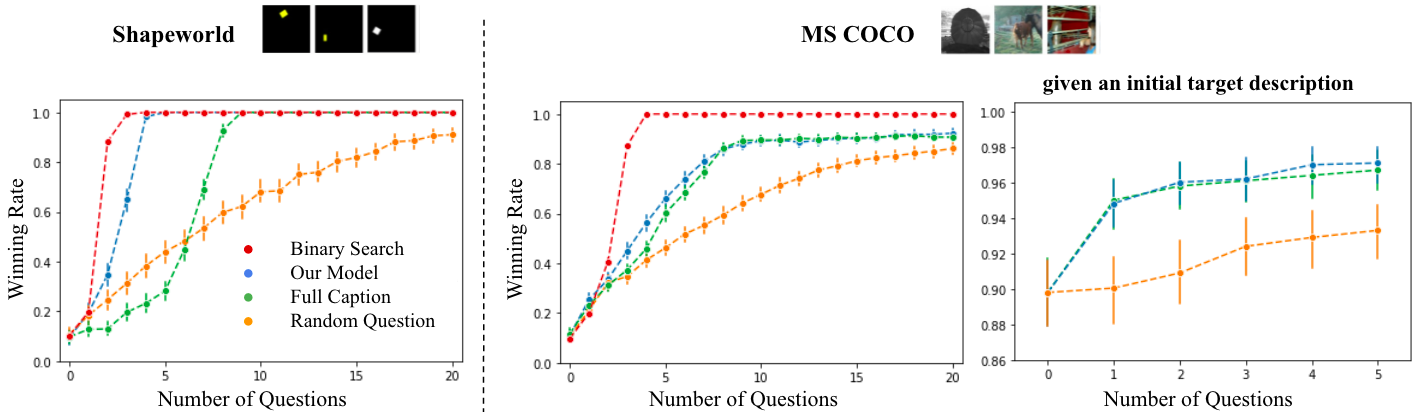} 
\end{center}
\caption{Winning rate curves for 10-image Shapeworld and MS COCO games. Error bars correspond to a 95\% confidence interval across games.}
\label{synthresults}
\end{figure*}

\subsection{Simulations on synthetic datasets}

Before deploying our model in interactions with human speakers, we examined its performance on synthetic datasets where we could carefully control the answerer.
We examine two domains: \textbf{Shapeworld} \cite{kuhnle2017}, a simple artificial dataset of images of random colored shapes paired with captions from a vocabulary of 15 words labeling the possible colors and shapes, and \textbf{MS COCO} \cite{lin2015}, a more naturalistic dataset containing images of everyday scenes paired with captions elicited by human annotators.
Because previous approaches have typically relied on closed-domain question-answer datasets or hand-built question semantics, they are incompatible with our ‘open domain’ setting. Instead, we compare our full model's performance against several model variants and strong, general-purpose search baselines: 
a \textbf{full caption} model which generates candidate questions from full image captions \emph{without decomposition}, comparable to a linear search checking one image at a time;
a \textbf{random question} model which selects questions randomly instead of using the expected information gain objective;
and, a \textbf{binary search} algorithm which serves as an upper-bound ``oracle,'' unfettered by the expressivity of real language, by randomly halving the set of potential target images with each step of the interaction rather than posing a natural language question. 
We evaluate these models on a total of 1,000 games sampled from each dataset using contexts of size $k=10$ images. 

For Shapeworld, we paired our questioner with an artificial answerer constructed to provide ground-truth answers to generated questions (Figure~\ref{synthresults}, left). 
Our proposed model outperforms the random baseline as well as the full caption model, which produces questions that are too specific to efficiently narrow the space of potential target images, while only slightly under-performing an upper-bound binary search algorithm.
These findings demonstrate the utility of having a question set of varying specificity (via decomposing full captions into NPs) as well as the expected information gain objective which adapts question selection to the model's current knowledge. 

\begin{table}[t!]
\centering
\begin{tabular}{l c} 
\hline
\textbf{Game Type} & \textbf{Entropy (95\% CI)}\\
\hline
Random context & 2.80 (2.76-2.83) \\
Split context & 2.60 (2.58-2.62) \\
Binary search & 2.32 (2.32-2.32) \\
\hline
\end{tabular}
\caption{Entropy after one question is asked for contexts with randomly sampled images or images split between two categories (10-image MS COCO games; binary search represents lower bound).
}
\label{tab:synthetic}
\end{table}

For MS COCO, we construct an artificial answerer that uses a simple heuristic, as ground-truth answers are not readily available for MS COCO. 
This answerer responds ``yes'' if a question is generated from the target image, and ``no'' otherwise (Figure~\ref{synthresults}, middle).
We again see that our model greatly outperforms a random questioner, but outperforms the full caption baseline to a lesser extent than we observed on Shapeworld. 
The larger gap between our model and binary search also indicates significant room for improvement.
One possible explanation for this gap is the difficulty of finding attributes which appropriately ``split'' a random set of natural images. 
To evaluate performance when a clear division of the image set is expressable in natural language, we created an alternative test set where we ensured that the 10 images in the context were balanced across two categories in COCO (i.e., five ``motorcycles'' and five ``baseballs''). 
We found that the model was indeed better able to divide the image set when we guaranteed that some high-level cut between the images existed (Table~\ref{tab:synthetic}).

When models were given an initial description of the target image before asking any questions (Figure~\ref{synthresults}, right), we see that questions are still useful -- improving accuracy by 6\% from the caption alone. 

\begin{table}[t] 
\centering
\begin{tabular}{l c} 
\hline
&\textbf{Winning Rate (95\% CI)}\\
\hline
Polar Questions & 72.6 (69.8, 75.4)\\
Polar and `What' \\
Questions &  75.1 (72.4, 77.8)\\
\hline
\end{tabular}
\caption{Winning rate after 20 questions are asked for 25-image MS COCO games played with synthetic answerers.}
\label{tab:what}
\end{table}

\textbf{Extension to wh-questions.} While our main results use polar questions exclusively, our framework has the potential to be extended to more general wh-questions. Using wh-movement rules we can derive questions from image captions that ask about more abstract properties of objects within images (e.g., given the caption “three men holding surfboards on a beach” we can straightforwardly derive questions like: “How many men are there?”, “Where are the men?”, or “What are the men holding?”). To illustrate this extension we provide preliminary results for simple `what' questions. We generate these questions by identifying instances of noun phrases followed by verb phrases in captions and transforming these into a set of `what' questions with single-word answers. We extract the noun (NN) and verb (VBG) from their respective phrases then produce questions of the form 'What is the <NN> <VBG>?'. To accommodate these questions in our model, we simply modified our answer classifier to produce a probability distribution over the entire vocabulary (rather than a binary yes-no). By incorporating what questions into our framework, we see an improvement of almost 3\% after 20 questions are asked (Table ~\ref{tab:what}). 

\subsection{Interactive human experiments}

\begin{table}[!t]
\centering
\begin{tabular}{l c} 
\hline
\textbf{} & \textbf{Winning Rate (95\% CI)}\\
\hline
Before Questions & 10.0 (5.6-14.4) \\
After Questions & 75.5 (67.8-83.1) \\
\hline
\textbf{Total Questions} & 6.39 (5.59-7.19)\\
\hline
\end{tabular}
\caption{Winning rate for 10-image MS COCO games played with human answerers.}
\label{tab:human1}
\end{table}

\begin{table}[!t]
\centering
\begin{tabular}{l c} 
\hline
\textbf{} & \textbf{Winning Rate (95\% CI)}\\
\hline
Before Questions & 58.5 (49.0-68.0) \\
After Questions & 75.0 (68.4-81.6) \\
\hline
\textbf{Total Questions} & 7.24 (5.53-8.95)\\
\hline
\end{tabular}
\caption{Winning rate for 25-image MS COCO games played with human answerers who give an initial target description.}
\label{tab:human2}
\end{table}

We ran two experiments to evaluate our question generation model in interactions with real human partners.
We recruited a total of 40 participants from Amazon Mechanical Turk to play 10 games each in which our model asked questions until the entropy of the belief distribution over images fell below 1.0 or until 20 questions were asked. Participants were prompted to give either a "yes", "no", or "N/A" response to each question.

In the first human experiment, games were sampled from the same 1,000 MS COCO games used for synthetic evaluation (Table~\ref{tab:human1}). We found that our question-asking model was able to successfully improve target selection accuracy when paired with a human answerer, suggesting that our model's questions are human interpretable and that human answers are effective for target selection.

Our second human experiment examines the more challenging case of asking ``clarification questions'' in a referential setting. 
In this experiment we used larger contexts of $k = 25$ images sampled from the MS COCO test set, and human participants were prompted to give a description of the target to initiate the interaction.
Our model formed (uncertain) beliefs based on this initial utterance and proceeded to ask clarification questions which we found improved by 16.5\% from the image description alone (see Table~\ref{tab:human2}).

\section{Conclusions}
We introduce a question generation framework capable of producing open-domain clarification questions. 
Instead of relying on specialized question-answer training data or pre-specified question meanings, our model uses a pretrained image captioner in conjunction with expected information gain to produce informative questions for unseen images.
We demonstrate the effectiveness of this method in a question-driven communication game with synthetic and human answerers.
We found it important to generate questions varying in specificity by decomposing captioner utterances into component noun phrases.
Having generated this set of potential questions, selecting based on estimated information gain yielded useful questions.
Without seeing question examples, our framework demonstrates a capacity for generating effective clarification questions. 

Future research should aim to generate more diverse question sets, allow for more expressive answers, and address abstract properties of objects within images. One approach, as demonstrated by our preliminary work with `what'-questions, would be to extend our framework to incorporate additional types of wh-questions. 
Integrating this clarification capacity more fully into collaborative, goal-directed dialog agents will allow them to engage in cooperative repair.

\section*{Acknowledgements}
This research was supported in part by the Office of Naval Research grant ONR MURI N00014-16-1-2007 and the Stanford HAI Hoffman--Yee project `Towards grounded, adaptive communication agents'.

\bibliography{anthology,custom}

\begin{thebibliography}{20}
\expandafter\ifx\csname natexlab\endcsname\relax\def\natexlab#1{#1}\fi

\bibitem[{Arkel et~al.(2020)Arkel, Woensdregt, Dingemanse, and
  Blokpoel}]{arkel2020}
Jacqueline~Van Arkel, Marieke Woensdregt, Mark Dingemanse, and Mark Blokpoel.
  2020.
\newblock A simple repair mechanism can alleviate computational demands of
  pragmatic reasoning: simulations and complexity analysis.
\newblock In \emph{Proceedings of the 24th Conference on Computational Natural
  Language Learning}, pages 177--194.

\bibitem[{Benotti and Blackburn(2021)}]{benotti2021}
Luciana Benotti and Patrick Blackburn. 2021.
\newblock Grounding as a collaborative process.
\newblock In \emph{Proceedings of the 16th Conference of the European Chapter
  of the Association for Computational Linguistics}, pages 515--531.

\bibitem[{Clark(1996)}]{clark1996}
Herbert~H. Clark. 1996.
\newblock \emph{Using Language}.
\newblock Cambridge University Press.

\bibitem[{Das et~al.(2017)Das, Kottur, Moura, Lee, and Batra}]{das2017}
Abhishek Das, Satwik Kottur, José M.~F. Moura, Stefan Lee, and Dhruv Batra.
  2017.
\newblock Learning cooperative visual dialog agents with deep reinforcement
  learning.
\newblock In \emph{International Conference on Computer Vision}.

\bibitem[{de~Vries et~al.(2017)de~Vries, Strub, Chandar, Pietquin, Larochelle,
  and Courville}]{vries2017}
Harm de~Vries, Florian Strub, Sarath Chandar, Olivier Pietquin, Hugo
  Larochelle, and Aaron Courville. 2017.
\newblock Guesswhat?! visual object discovery through multi-modal dialogue.
\newblock In \emph{Proceedings of the IEEE Conference on Computer Vision and
  Pattern Recognition}, pages 5503--5512.

\bibitem[{Kitaev et~al.(2019)Kitaev, Cao, and Klein}]{kitaev2019}
Nikita Kitaev, Steven Cao, and Dan Klein. 2019.
\newblock Multilingual constituency parsing with self-attention and
  pre-training.
\newblock In \emph{Proceedings of the 57th Annual Meeting of the Association
  for Computational Linguistics}, pages 3499--3505.

\bibitem[{Kovashka and Grauman(2013)}]{kovashka2013}
Adriana Kovashka and Kristen Grauman. 2013.
\newblock Attribute pivots for guiding relevance feedback in image search.
\newblock In \emph{International Conference on Computer Vision}.

\bibitem[{Kuhnle and Copestake(2017)}]{kuhnle2017}
Alexander Kuhnle and Ann Copestake. 2017.
\newblock Shapeworld - a new test methodology for multimodal language
  understanding.
\newblock \emph{arXiv preprint arXiv:1704.04517}.

\bibitem[{Kumar and Black(2020)}]{kumar2020}
Vaibhav Kumar and Alan~W. Black. 2020.
\newblock Clarq: A large-scale and diverse dataset for clarification question
  generation.
\newblock In \emph{Proceedings of the 58th Annual Meeting of the Association
  for Computational Linguistics}, pages 7296--7301.

\bibitem[{Lee et~al.(2018)Lee, Heo, and Zhang}]{lee2018}
Sang-Woo Lee, Yu-Jung Heo, and Byoung-Tak Zhang. 2018.
\newblock Answerer in questioner’s mind: Information theoretic approach to
  goal-oriented visual dialog.
\newblock In \emph{Proceedings of the 32nd Conference on Neural Information
  Processing Systems}, pages 2579--2589.

\bibitem[{Lewis(1969)}]{lewis1969}
David~K. Lewis. 1969.
\newblock \emph{Convention: A Philosophical Study}.
\newblock Harvard University Press.

\bibitem[{Lin et~al.(2015)Lin, Maire, Belongie, Bourdev, Girshick, Hays,
  Perona, Deva~Ramanan, and Dollár}]{lin2015}
Tsung-Yi Lin, Michael Maire, Serge Belongie, Lubomir Bourdev, Ross Girshick,
  James Hays, Pietro Perona, C.~Lawrence~Zitnick Deva~Ramanan, and Piotr
  Dollár. 2015.
\newblock Microsoft coco: Common objects in context.
\newblock In \emph{European Conference on Computer Vision}, pages 740--755.

\bibitem[{Misra et~al.(2018)Misra, Girshick, Fergus, Hebert, Gupta, and Van
  Der~Maaten}]{misra2018}
Ishan Misra, Ross Girshick, Rob Fergus, Martial Hebert, Abhinav Gupta, and
  Laurens Van Der~Maaten. 2018.
\newblock Learning by asking questions.
\newblock In \emph{Proceedings of the IEEE Conference on Computer Vision and
  Pattern Recognition}, pages 11--20.

\bibitem[{Purver et~al.(2002)Purver, Ginzburg, and Healey}]{purver2001}
Matthew Purver, Jonathan Ginzburg, and Patrick Healey. 2002.
\newblock On the means for clarification in dialogue.
\newblock In Jan van Kuppevelt and Ronnie~W. Smith, editors, \emph{Current and
  New Directions in Discourse and Dialogue}, pages 235--255. Kluwer Academic
  Publishers.

\bibitem[{Rao and Daumé~III(2018)}]{rao2018}
Sudha Rao and Hal Daumé~III. 2018.
\newblock Learning to ask good questions: Ranking clarification questions using
  neural expected value of perfect information.
\newblock In \emph{Proceedings of the 56th Annual Meeting of the Association
  for Computational Linguistics}, pages 2737–--2746.

\bibitem[{Rothe et~al.(2017)Rothe, Lake, and Gureckis}]{rothe2017}
Anselm Rothe, Brenden~M. Lake, and Todd Gureckis. 2017.
\newblock Question asking as program generation.
\newblock In \emph{Proceedings of the 31st Conference on Neural Information
  Processing Systems}, pages 1046--1055.

\bibitem[{Stivers(2010)}]{stivers2010}
Tanya Stivers. 2010.
\newblock An overview of the question–response system in american english
  conversation.
\newblock \emph{Journal of Pragmatics}, 42(10):2772--2781.

\bibitem[{Wang and Lake(2019)}]{wang2020}
Ziyun Wang and Brenden~M. Lake. 2019.
\newblock Modeling question asking using neural program generation.
\newblock \emph{arXiv preprint arXiv:1704.04517}.

\bibitem[{Yao et~al.(2018)Yao, Zhang, Luo, Tao, and Wu}]{yao2018}
Kaichun Yao, Libo Zhang, Tiejian Luo, Lili Tao, and Yanjun Wu. 2018.
\newblock Teaching machines to ask questions.
\newblock In \emph{International Joint Conferences on Artificial Intelligence},
  pages 4546--4552.

\bibitem[{Yu et~al.(2020)Yu, Chen, Wang, Lei, and Artzi}]{yu2020}
Lili Yu, Howard Chen, Sida Wang, Tao Lei, and Yoav Artzi. 2020.
\newblock Interactive classification by asking informative questions.
\newblock In \emph{Proceedings of the 58th Annual Meeting of the Association
  for Computational Linguistics}, pages 2664--–2680.

\end{thebibliography}
\bibliographystyle{acl_natbib}

\setcounter{figure}{0}
\setcounter{table}{0}

\newpage

\section*{Appendix A: Candidate question details}

We used a transformer-based image captioning architecture\footnote{Pre-trained model  \href{https://github.com/krasserm/fairseq-image-captioning}{https://github.com/krasserm/fairseq-image-captioning}} pre-trained on the MS COCO dataset, and a greedy search algorithm to generate one caption per image. See Table ~\ref{examples} for examples. 

\section*{Appendix B: Experiment details}

For ShapeWorld, a set of 1,000,000 images and their captions (which can include the shape and/or color of the object depicted) was used to train a Shapeworld-specific image captioner and answerer model.

For MS COCO, the image captioner and answerer model were trained on the Karpathy splits which allocate 155,000 samples for training and 5,000 images for validation and testing each.
The images used in our games were randomly drawn from the test set. We used a vocabulary size of 9,808 words.

\section*{Appendix C: Model and baseline outputs} 

\begin{table}[h]
\centering
\includegraphics[width=7.5cm]{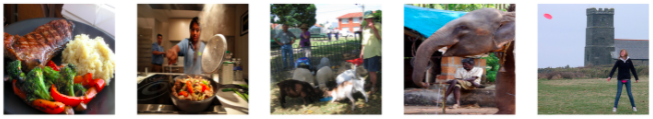}
\begin{tabular}{l l} 
\hline
\textbf{Our Model} & Is there food?\\
\textbf{Full Caption} & Is there a woman in a \\
&  kitchen cooking food\\
&  on a stove?\\
\textbf{Random Question} & Is there a herd?\\
\hline
\\
\end{tabular}
\includegraphics[width=7.5cm]{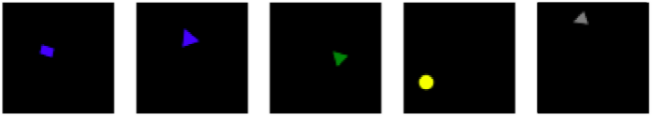}
\begin{tabular}{l c} 
\hline
\textbf{Our Model} & Is there a triangle?\\
\textbf{Full Caption} & Is there a blue square?\\
\textbf{Random Question} & Is there a gray circle?\\
\hline
\end{tabular}
\caption{Questions selected by our model and baselines for 10-image MSCOCO (top) and Shapeworld (bottom) games.}
\label{modelout}
\end{table} 

Examples outputs for our model and each of the baselines presented are given in Table ~\ref{modelout}. Binary search is not included because we do not pose natural language questions, and instead randomly split the image set in half with each "question".

\section*{Appendix D: Accuracy-efficiency tradeoff}

\begin{figure}[!h]
\begin{center}
\includegraphics[width=6.25cm]{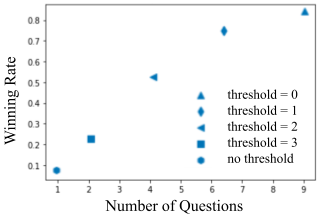}
\end{center}
\caption{Winning rate vs. the number of questions asked at different entropy thresholds for 10-image MS COCO games played with human answerers.}
\label{entropy}
\end{figure}

In our human experiments our model asked questions until the entropy of the belief distribution over images fell below 1.0. However, this threshold value can be raised or lowered to produce a higher communication accuracy or a lower number of questions. Figure ~\ref{entropy} shows the accuracy-efficiency tradeoff at different entropy threshold values.

\section*{Appendix E: Human-selected questions} 

\begin{table}[h]
\centering
\begin{tabular}{l c} 
\hline
&\textbf{Entropy (95\% CI)}\\
\hline
Model & 1.06 (0.99, 1.12)\\
Human &  1.32 (1.25, 1.38)\\
\hline
\end{tabular}
\caption{Entropy after one question is asked for human selected and model-selected questions (6-image MS COCO games).}
\label{tab:humanselect}
\end{table}

We asked 38 participants from Amazon Mechanical Turk to rerank ten sets of five questions by their informativity. Participants were shown a set of six images and prompted with a target image description then asked to rank their set of questions according to which they would be most likely to ask. In Table ~\ref{tab:humanselect} we show the entropy after asking human-selected and model-selected questions (given the same image set, initial description, and question set). This comparison may not be entirely fair as the human's and model's beliefs are not fully aligned, and what may be the most informative question for a human may not be the most informative question for the model. However, we do see that the model-selected questions ultimately produce a lower entropy than human-selected questions.

\begin{table*}
\begin{center}
\includegraphics[width=16cm]{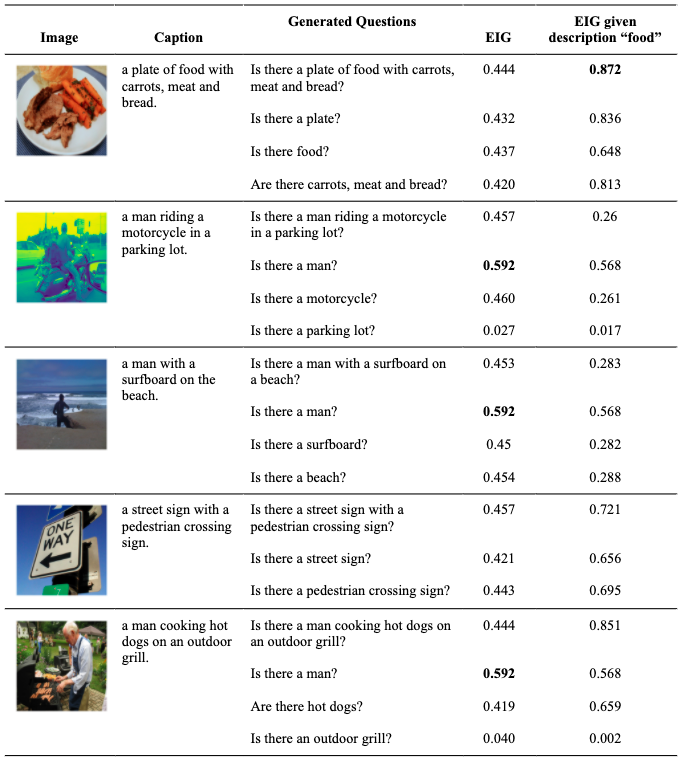}
\end{center}
\caption{Example candidate questions generated for MS COCO images and their expected information gain (EIG) with and without the initial target image description "food".}
\label{examples}
\end{table*} 

\end{document}